\documentclass[letterpaper, 10 pt, conference]{ieeeconf}  
                
\usepackage{graphicx}
\usepackage{multirow}
\usepackage{algorithm}
\usepackage{soul}
\usepackage{comment}
\usepackage[noend]{algpseudocode}
\usepackage{afterpage}
\usepackage{array}
\usepackage{amsmath}
\usepackage{amsmath} 
\usepackage{balance}
\usepackage{makecell}
\usepackage{hyperref}
\usepackage{titlesec}

\IEEEoverridecommandlockouts                              
\newcolumntype{C}{>{\centering \arraybackslash}m{0.1\textwidth}}

\title{\LARGE \bf

SwarmVLM: VLM-Guided Impedance Control for Autonomous Navigation of Heterogeneous Robots in Dynamic Warehousing} 

\author{Malaika Zafar, Roohan Ahmed Khan\textsuperscript{*}, Faryal Batool\textsuperscript{*}, Yasheerah Yaqoot\textsuperscript{*}, Ziang Guo, \\Mikhail Litvinov, Aleksey Fedoseev, and Dzmitry Tsetserukou%
\thanks{The authors are with the Intelligent Space Robotics Laboratory, Skolkovo Institute of Science and Technology, Moscow, Russia. 
\tt \{malaika.zafar, roohan.khan, faryal.batool, yasheerah.yaqoot, ziang.guo, mikhail.litvinov, aleksey.fedoseev, d.tsetserukou\}@skoltech.ru}
\thanks{*These authors contributed equally to this work.}
}

\begin{document}
\maketitle
\thispagestyle{empty}
\pagestyle{empty}
\begin{abstract}


With the growing demand for efficient logistics, unmanned aerial vehicles (UAVs) are increasingly being paired with automated guided vehicles (AGVs). While UAVs offer the ability to navigate through dense environments and varying altitudes, they are limited by battery life, payload capacity, and flight duration, necessitating coordinated ground support.

Focusing on heterogeneous navigation, \textbf{SwarmVLM} addresses these limitations by enabling semantic collaboration between UAVs and ground robots through impedance control. The system leverages the Vision Language Model (VLM) and the Retrieval-Augmented Generation (RAG) to adjust impedance control parameters in response to environmental changes. In this framework, the UAV acts as a leader using Artificial Potential Field (APF) planning for real-time navigation, while the ground robot follows via virtual impedance links with adaptive link topology to avoid collisions with short obstacles. 

The system demonstrated a 92\% success rate across 12 real-world trials. Under optimal lighting conditions, the VLM-RAG framework achieved 80\% accuracy in object detection and selection of impedance parameters. The mobile robot prioritized short obstacle avoidance, occasionally resulting in a lateral deviation of up to 50 cm from the UAV path, which showcases safe navigation in a cluttered setting. 

Video of SwarmVLM: \href{https://youtu.be/IdlUhQfz8w0}{https://youtu.be/IdlUhQfz8w0}
\end{abstract}


\begin{figure}[htbp]
\centering
\includegraphics[width=1\linewidth]{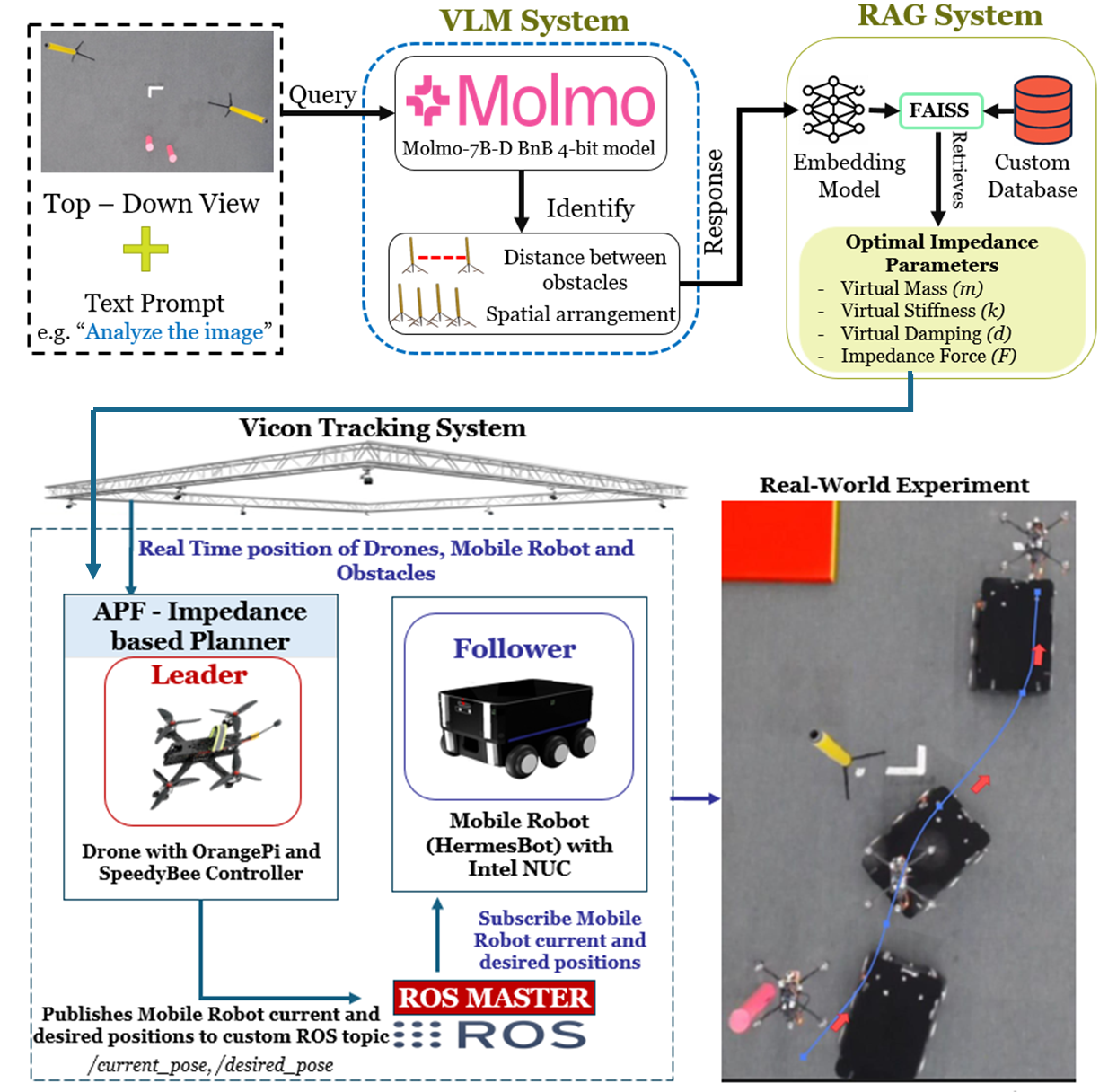}
\caption{ Framework for adaptive swarm navigation. The system processes a top-down view of the environment through the VLM-RAG system to determine impedance parameters based on the arrangements and number of obstacles in the environment. Experiments were also performed in the real-world environment to evaluate the system’s robustness and generalization capabilities. 
}
\vspace{-5mm} 
\label{fig:main_image}
\end{figure}

\section{Introduction}
Advancements in automation and artificial intelligence have significantly improved logistics and warehouse management by enabling autonomous systems that reduce human error and improve operational efficiency~\cite{Lopez_2023}. Automated guided vehicles (AGVs) and mobile robots have traditionally dominated indoor and last-mile delivery due to their reliable localization and high payload capacity~\cite{Motroni_2024}.
Unmanned aerial vehicles (UAVs), in contrast, offer fast and flexible movement in three-dimensional space, making them ideal for navigating over obstacles and through constrained environments. For instance, Cristiani et al.~\cite{Cristiani_2020} proposed a mini-drone swarm for inventory tracking. However, UAVs are limited by payload, flight time, and sensing range. Integrating UAVs with ground robots enables a hybrid system that combines aerial agility with ground-level stability and endurance.

To enable such heterogeneous cooperation, we propose a leader-follower framework called \textbf{SwarmVLM}, where the drone leads using an Artificial Potential Field (APF)-based planner, and the mobile robot follows via virtual impedance linkages~\cite{Hogan_1984}. The mobile robot also forms temporary impedance links with short obstacles undetectable by the drone, while a custom PID controller minimizes its path deviation. This architecture ensures reliable operation in dense and dynamic environments.
To further enhance adaptability, we integrate a Vision-Language Model (VLM) with a Retrieval-Augmented Generation (RAG) framework. This \textbf{VLM-RAG} module interprets top-down environmental views and retrieves context-specific impedance parameters from a scenario database. It enables both agents to adjust their impedance settings based on obstacle arrangement and proximity, ensuring safe and intelligent navigation.

SwarmVLM introduces several key contributions:
\begin{itemize}
    \item An impedance-based coordination mechanism for drone–robot cooperation in cluttered environments.
    \item Integration of the VLM-RAG framework for understanding environmental configuration and adaptive impedance tuning.
    \item Real-world validation of the system in dynamic indoor environments.
\end{itemize}
\section{Related Work}
Logistics and warehouse management have become critical components of modern supply chains, driving the demand for more efficient and autonomous delivery solutions. In response, both homogeneous and heterogeneous swarms of robotic agents have been explored to meet these challenges.

Managing large UAV swarms, however, presents significant difficulties in coordination, stability, and real-time decision-making. To address this, several studies have introduced human-in-the-loop control approaches. For instance, Abdi et al.~\cite{Abdi_2023} leveraged EMG signals to enable gesture-based swarm control via muscle activity, while Khen et al.~\cite{Khen_2023} combined gesture recognition with machine learning for intuitive drone manipulation through natural human movements.

Despite these advances, achieving fully autonomous and reliable swarm navigation in complex and dynamic environments remains a major challenge. Many existing UAV systems still struggle with robust perception, real-time adaptation, and decentralized coordination without human oversight. Among the most widely used methods for autonomous navigation is the Artificial Potential Field (APF) technique, appreciated for its simplicity and reactive obstacle avoidance capabilities. Batinovic et al.~\cite{APF_LiDAR} integrated APF with LiDAR for real-time path planning in unstructured environments, demonstrating its suitability for navigating dense and dynamic spaces. Yu et al.~\cite{Yu_2023} proposed a distributed algorithm that couples APF with virtual leader formation and a switching communication topology to ensure robust swarm behavior.

To improve swarm cohesion and control, impedance-based strategies have been introduced. Tsykunov et al.~\cite{Tsykunov_2019} developed SwarmTouch, which utilized virtual impedance links~\cite{Hogan_1984} for drone swarm coordination. Building on this, Fedoseev et al.~\cite{Fedoseev_2022} analyzed the influence of impedance topologies on swarm stability. Khan et al.~\cite{SwarmPath} later combined APF-based leader planning with impedance-based formation control, but their approach remained limited to static environments and did not address challenges like energy efficiency or long-term operation.

To overcome the limitations of aerial-only or ground-only systems, research has increasingly focused on heterogeneous swarm architectures. Darush et al.~\cite{SwarmGear} employed virtual impedance links between a leader octocopter and micro-drones, while ~\cite{Karaf_2023} used reinforcement learning to facilitate docking, transport, and in-air recharging. Chen et al.~\cite{Chen_2025} proposed a UAV-AGV system for collaborative exploration in hazardous areas, highlighting the advantages of combining agents with complementary capabilities to enhance flexibility and robustness.

Ground robots also play a crucial role in warehouse logistics. Malopolski et al.~\cite{GroundRobot} introduced an autonomous mobile robot equipped with a hybrid drive system for navigating flat and rail surfaces, and an integrated elevator for vertical mobility. While effective in structured spaces, such systems lack aerial reach and adaptability in cluttered or hard-to-access environments.

Further demonstrating the benefits of aerial-ground collaboration, Salas et al.~\cite{Salas_2021} presented a UAV-AGV system for search and rescue. Their UAV provided aerial pathfinding using a monocular camera, while the ground robot performed close-range inspection and mapping, showing the value of layered autonomy in unstructured terrains.

Parallel to these developments, vision-based learning models, particularly Vision Transformers (ViTs), have shown promise in robotic perception and decision-making. Dosovitskiy et al.~\cite{visiontransformerdosovitskiy2021imageworth16x16words} demonstrated the effectiveness of transformers for image recognition, paving the way for their use in aerial robotics for tasks like object detection and spatial understanding. More recently, Brohan et al.~\cite{brohan2023rt1roboticstransformerrealworld} introduced RT-1, a task-conditioned transformer capable of generalizing across diverse robotic control tasks in real-world environments.

Multimodal transformers such as Molmo-7B-D~\cite{molmo_quantized} integrate visual, spatial, and linguistic inputs, enabling rich contextual awareness for autonomous systems. These architectures are especially useful for coordination and decision-making in cluttered or uncertain environments. For instance, FlockGPT~\cite{FlockGPT} introduced a generative AI interface that allows users to control drone flocks using natural language, achieving high accuracy and strong usability for dynamic shape formation.

Building on these foundations, our research proposes a novel heterogeneous swarm system that combines APF-based aerial navigation, impedance-guided ground mobility, and a VLM-RAG-powered framework for adaptive, context-aware behavior. Inspired by prior work on HetSwarm~\cite{HetSwarm} and ImpedanceGPT~\cite{impedancegpt}, this study introduces a new agile and safe path planning solution for drone–robot teams operating in dynamic and cluttered environments.
\begin{figure*}[t!]
\centering
\includegraphics[width=0.8\linewidth]{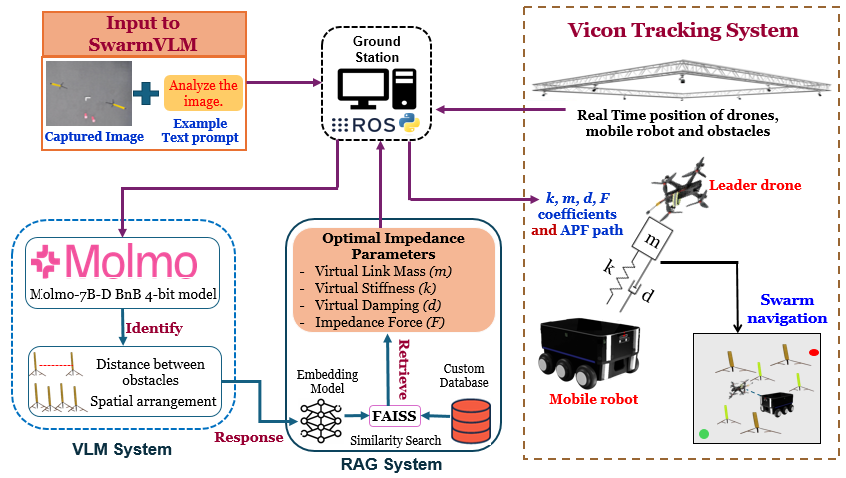}
\caption{System architecture of the proposed \textit{SwarmVLM}, integrating a VLM-RAG module for impedance estimation with a high-level control system that combines APF-based path planning and impedance control for heterogeneous navigation.}
\label{system arch}
\vspace{-5mm}
\end{figure*}
\section{SwarmVLM Technology}
The proposed methodology, illustrated in Fig.~\ref{system arch}, consists of two primary components: a VLM-RAG system for estimating impedance parameters and a high-level control framework that integrates APF planner with impedance-based coordination.
The system enables collaboration between a drone and a mobile robot in dynamic environments. The drone uses the APF planner to compute and update its trajectory in real-time, focusing on global navigation and obstacle avoidance. The mobile robot follows this trajectory through virtual impedance links, which help maintain formation and allow local navigation around small obstacles not perceived by the drone. Additionally, the robot can act as a mobile landing or recharging platform for the drone.

The methodology was first validated in the Gym PyBullet \cite{gympybullet} simulation environment using custom PID controllers for both agents to ensure accurate and coordinated path following. Real-world experiments were then conducted, demonstrating effective drone–robot collaboration in cluttered environments. Communication between the drone and the mobile robot is handled via ROS, enabling synchronized behavior through a shared information framework.

\subsection{Artificial Potential Fields for Global Path Generation}

In order for the leader drone to navigate efficiently around the obstacles while setting the path toward the goal, we applied the APF planning algorithm \cite{APF}. This algorithm  allows the UAV to continuously update its trajectory in response to shifting obstacles, enabling robust navigation in cluttered and dynamic environments. The equations for the APF planner are as follows \cite{SwarmPath}: 
\begin{equation}
F_{\text{total}} = F_{\text{attraction}} + F_{\text{repulsion}} \label{eq:total_force},
\end{equation}

where

\begin{align*}
F_{\text{attraction}}(d_{\text{g}}) &= k_{\text{att}} \cdot d_{\text{g}}, \\
F_{\text{repulsion}}(d_{\text{o}}) &= 
\begin{cases} 
0 & \text{if } d_{\text{o}} > d_{\text{safe}} \\
k_{\text{rep}} \cdot \left( \frac{1}{d_{\text{o}}} - \frac{1}{d_{\text{safe}}} \right) & \text{if } d_{\text{o}} \leq d_{\text{safe}},
\end{cases}
\end{align*}
where $d_{\text{g}}$ and $d_{\text{o}}$ are the distances from the drone to the goal and to the obstacle, respectively, $k_{\text{att}}$ and $k_{\text{rep}}$ are the attraction and repulsion coefficients, respectively.

\subsection{Impedance Controller}
To enable smooth coordination, a virtual impedance controller couples the drone and the mobile robot. The mobile robot acts as a follower, linked to the drone’s APF-based trajectory through a mass spring damper system that ensures stable formation tracking. This dynamic coupling is governed by:

\begin{equation}
m\Delta\ddot{x} + d\Delta\dot{x} + k\Delta x = F_{\text{ext}}(t),
\label{eq:dynamic_equation}
\end{equation}
where $\Delta x$, $\Delta\dot{x}$, and $\Delta\ddot{x}$ represent deviations in position, velocity, and acceleration from the desired state; \(m\), \(d\), and \(k\) are virtual mass, damping, and stiffness; and \(F_{\text{ext}}(t)\) is the virtual force from the drone.

To handle short obstacles undetected by the drone, the mobile robot temporarily disengages from the drone and forms local impedance links with the obstacles. The resulting repulsive displacement is given by:

\begin{equation}
\Delta x_{\text{robot}} = k_{\text{impF}} \cdot r_{\text{imp}},
\label{eq:deflection_eqn}
\end{equation}
where \(k_{\text{impF}}\) is the velocity-dependent force coefficient and \(r_{\text{imp}}\) defines the influence radius of the obstacle. This mechanism enables timely, collision-free navigation in cluttered environments. This equation ensures a collision-free trajectory by applying a repulsive displacement proportional to the obstacle’s influence radius $r_{imp}$, redirecting the robot away from potential collisions. The coefficient \(k_{\text{impF}}\) adjusts the strength of this deflection based on the robot's velocity, ensuring timely and stable avoidance.


\subsection{VLM-RAG System}
The VLM-RAG system processes a top-down visual view of the environment using the Molmo-7B-D BnB 4-bit model to extract key features such as the number and spatial distribution of obstacles. These features are converted into vector representations and passed to the Retrieval-Augmented Generation (RAG) system, which retrieves optimal impedance control parameters — namely, virtual mass ($m$), virtual stiffness ($k$), virtual damping ($d$), and the impedance force coefficient ($F$). Communication between the VLM-RAG module, running on the server, and the heterogeneous swarm system is handled through the ROS framework.

\subsubsection{Integrating VLM for Obstacle Identification and Spatial Analysis}

The Vision-Language Model (VLM) uses visual input from a ceiling-mounted camera to semantically analyze the environment by detecting, and localizing obstacles. For efficient real-time performance, the lightweight Molmo-7B-D BnB 4-bit model~\cite{molmo_quantized} is employed. The extracted obstacle information is then forwarded to the RAG system, which uses the spatial configuration to retrieve suitable impedance control parameters, thereby enhancing swarm navigation in both static and dynamic environments.

\subsubsection{Custom Database for multiple environmental cases}
A custom database of six environmental scenarios was created to support the RAG system. Each scenario represents an indoor space with different obstacle quantities and spatial arrangements. The database stores empirically derived optimal impedance parameters obtained through simulation experiments using the APF planner for drone navigation. Table~\ref{tab:impedance_params_cases} summarizes the parameters for all six cases.

\begin{table}[h!]
    \centering
    \caption{Database contains the optimal impedance parameters, including virtual mass (m), virtual stiffness (k), virtual damping (d), and impedance or deflection force coefficient (F), for six different cases
    }
    \renewcommand{\arraystretch}{1.2} 
    \begin{tabular}{|c|c|c|c|c|}
        \hline
        \textbf{Parameter} & \textbf{$m$ (kg) } & \textbf{$k$ (N/m)} & \textbf{$d$ (N$\cdot$s/m)} & \textbf{$F_{\text{coeff}}$} \\
        \hline
        Case I      & 1.0 & 5.0 & 2.5 & 0.68 \\
        \hline
        Case II      & 1.5 & 3.0 & 3.5 & 0.35  \\
        \hline
        Case III      & 1.2 & 3.5 & 3.0 & 0.45  \\
        \hline
        Case IV      & 1.3 & 3.4 & 3.6 & 0.45 \\
        \hline
        Case V      & 1.4 & 3.3 & 3.7 & 0.15  \\
        \hline
        Case VI      & 1.2 & 3.8 & 4.0 & 0.65 \\
        \hline
    \end{tabular}
    \label{tab:impedance_params_cases}
    \vspace{-8pt}
    
\end{table}

\subsubsection{Retrieval-Augmented Generation (RAG) for Impedance Parameter Generation}
Retrieval-Augmented Generation (RAG) systems typically enhance Large Language Models (LLMs) by integrating external knowledge retrieval into the generation process. In this work, a naïve RAG implementation is used. It employs a sentence transformer to generate vector embeddings from textual queries and utilizes Facebook AI Similarity Search (FAISS) for fast and efficient nearest-neighbor retrieval. An exact nearest-neighbor search algorithm is adopted to ensure precise matching.
The RAG pipeline processes the textual output of the VLM, embeds it as a query vector, and performs a similarity search in the database of impedance parameters using Euclidean distance given by: 
\begin{equation}
d(X, Y) = \sqrt{\sum_{i=1}^{n} (X_i - Y_i)^2,}
\end{equation}
where 
\( X = (X_1, X_2, ..., X_n) \) is the query embedding,  
\( Y = (Y_1, Y_2, ..., Y_n) \) is the stored embedding,  
\( n = 384 \) is the embedding dimension.

The closest matching case is returned, and its associated impedance parameters are used to adjust the behavior of the swarm according to the current environmental context.

\section{Experimental Setup in Simulation Environment}
\begin{figure}[H]
\centering
\includegraphics[width=0.95\linewidth]{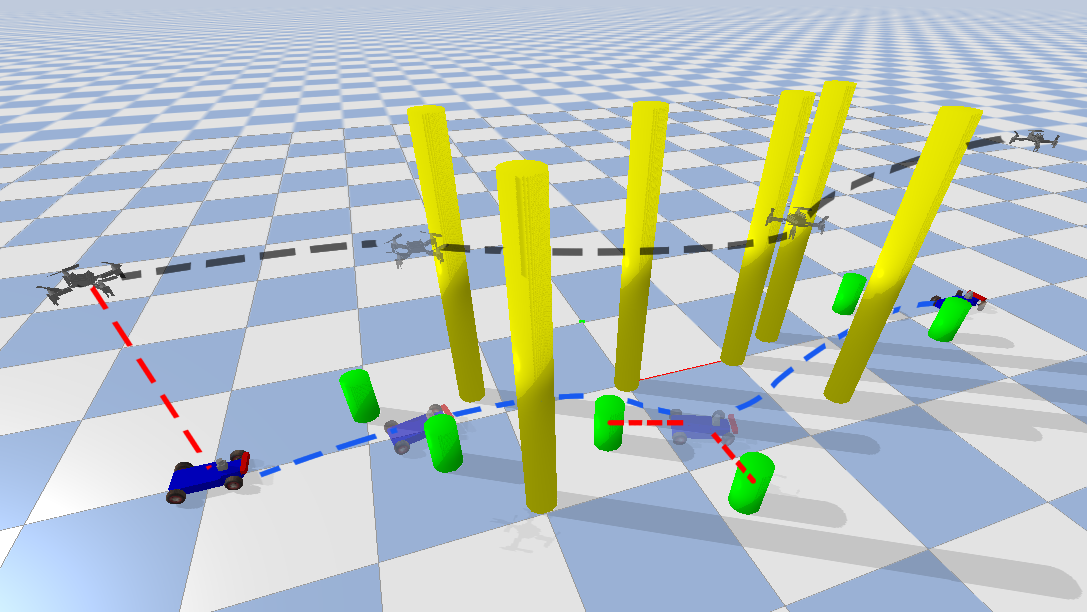} 
\caption{Experimental setup in the Gym Pybullet environment showing the trajectory of the drone and mobile robot under a dense environment. Where the red line shows the impedance linkages and the black and blue lines show the drone and mobile robot paths, respectively.}
\label{exp_gym}
\vspace{-0.2cm}

\end{figure}

A custom simulation environment was developed using the Gym-PyBullet framework (Fig.~\ref{exp_gym}) to model the dynamics of the drone and mobile robot. The setup supports real-time visualization and allows dynamic repositioning of obstacles for flexible testing. The drone navigates using an Artificial Potential Field (APF) planner, while the mobile robot follows the drone's trajectory through impedance-based coordination.

\section{Real-World Experimental Setup}
The real-world setup consists of a drone and a mobile robot communicating via ROS over a shared Wi-Fi network. The drone operates as the ROS master using an onboard Orange Pi, while the mobile robot functions as a ROS slave running on an Intel NUC. Initially operating independently, both agents are synchronized through ROS for collaborative operation.

The drone continuously publishes its planned target position and the mobile robot’s current position (as tracked by a VICON motion capture system). Upon synchronization, the mobile robot subscribes to this data and uses a PID controller to compute velocity commands, enabling it to follow the drone’s trajectory while maintaining formation and avoiding local obstacles. A total of 12 experiments were conducted with varying environmental configurations by altering obstacle positions.

\subsection{Experimental Result}
Trajectories from three of the experiments are shown: two in static environments, and Case III evaluated under both static and dynamic conditions.

\subsubsection{Results in static environment}
\begin{figure}[htbp]
\centering
\includegraphics[width=1\linewidth]{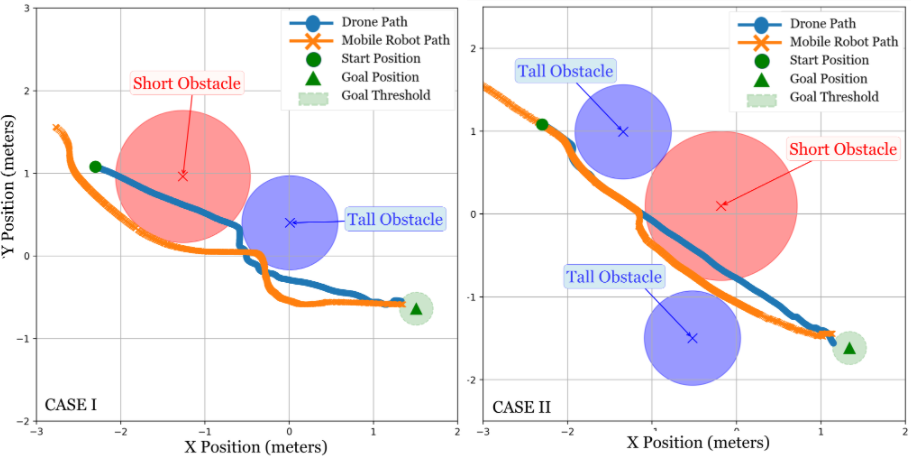} 
\caption{Results in static environment. CASE I: One tall obstacle, one short obstacle, CASE II: two tall obstacles, one short obstacle.}
\label{exp_1}
\vspace{-0.2cm}
\end{figure}

Fig.~\ref{exp_1} demonstrates how the drone leads the mobile robot through virtual impedance-based coupling while successfully avoiding tall obstacles. The mobile robot actively avoids short ground obstacles, which are not perceived or avoided by the drone since they do not interfere with its flight path.

\subsubsection{Results in dynamic environment}
\begin{figure}[htbp]
\centering
\includegraphics[width=1\linewidth]{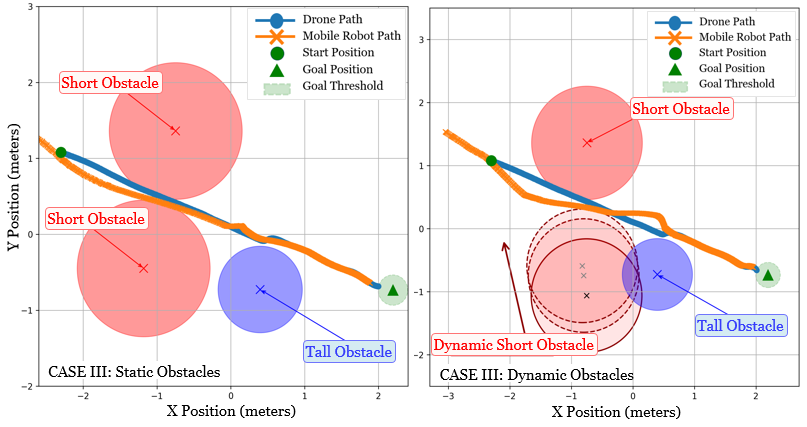} 
\caption{Results in a dynamic environment. CASE III: One Tall Obstacle – Two Short Obstacles.}
\label{exp_2}
\vspace{-0.2cm}
\end{figure}

In Case III (Fig.~\ref{exp_2}), one of the short obstacles was made dynamic to simulate a real-world scenario. As the responsibility for avoiding short obstacles lies solely with the mobile robot, the figure shows that while the drone follows the same planned trajectory in both static and dynamic scenarios, the mobile robot adjusts its path in real time to avoid the approaching obstacle.

\subsubsection{Deviation of mobile robot path from drone path}
While the drone demonstrates minimal deviation from its planned path due to real-time APF-based planning, the mobile robot deviates to avoid short obstacles that are not in the path of the drone, especially in cluttered environments. These deviations highlight the mobile robot’s role in ensuring collision avoidance while maintaining formation. The quantitative deviations are summarized in Table~\ref{tab:trajectory_deviation_horizontal}.

\begin{table}[h!]
    \centering
    \caption{Mobile Robot Path Offset due to Short Obstacle Avoidance }
    \renewcommand{\arraystretch}{2.2}
    \begin{tabular}{|c|c|c|c|c|}
        \hline
        \textbf{Case} & \textbf{I} & \textbf{II} & \makecell{\textbf{III}\\\textbf{(Static)}} & \makecell{\textbf{IV}\\\textbf{(Dynamic)}} \\
        \hline
        \textbf{Mobile Robot Deviation (m)} & 0.45 & 0.45 & 0.45 & 0.43 \\
        \hline
    \end{tabular}
    \label{tab:trajectory_deviation_horizontal}
    \vspace{-8pt}
    
\end{table}

\subsubsection{Velocity Distribution along the trajectory}
\begin{figure}[htbp]
\centering
\includegraphics[width=1\linewidth]{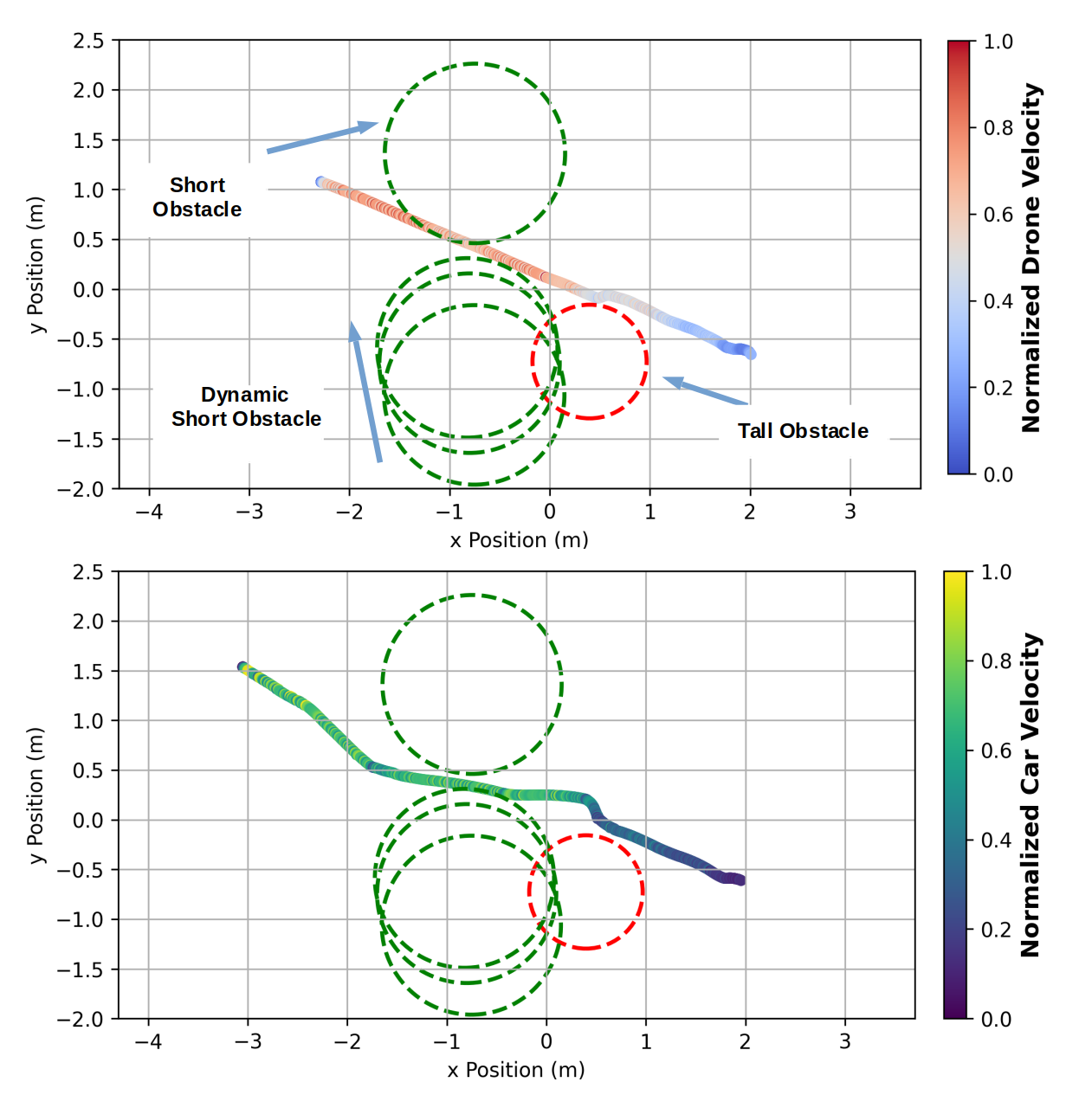} 
\caption{Velocity profile along the trajectory in real world scenarios.}
\label{velocity_real}
\end{figure}

The velocity profiles in Fig.~\ref{velocity_real} capture the real-world motion behavior of both agents. The drone maintains a relatively high and consistent velocity, with slight accelerations near tall obstacles followed by gradual deceleration. The mobile robot exhibits a more variable profile, accelerating when simultaneously following the drone and avoiding short obstacles, and decelerating as it reaches the final target.

\subsection{Trajectory Analysis Across Experimental Cases}
Table~\ref{tab:trajectory_comparison} shows the trajectory lengths of the drone and mobile robot. The robot consistently takes a longer path due to short obstacle avoidance, while the drone flies directly from them. Case III highlights performance in both static and dynamic settings, reflecting system adaptability.

\begin{table}[h!]
    \centering
    \caption{Trajectory Length Comparison Between Drone and Mobile Robot}
    \label{tab:trajectory_comparison}
    \renewcommand{\arraystretch}{2}
    \begin{tabular}{|c|c|c|}
        \hline
       \textbf{Case} & \makecell{\textbf{Drone}\\\textbf{trajectory (m)}} & \makecell{\textbf{Mobile Robot}\\\textbf{trajectory (m)}} \\
        \hline
        Case I & 4.272 & 5.593 \\
        \hline
        Case II & 4.492 & 5.822 \\
        \hline
        Case III (Static) & 4.684 & 5.634 \\
        \hline
        CASE III (Dynamic) & 4.672 & 5.710 \\
        \hline
    \end{tabular}
    \vspace{-9.5pt}
\end{table}

Experiments were conducted under varying conditions. Out of 12 trials, 11 of them succeeded. One failure resulted from synchronization issues between both the agents and limited flight area due to the large mobile robot's size. The success rate was computed to be 92\%. This high success rate demonstrates the robustness of the system and its capability to adapt to dynamic and diverse real-world conditions.

\begin{figure}[htbp]
\centering
\includegraphics[width=1\linewidth]{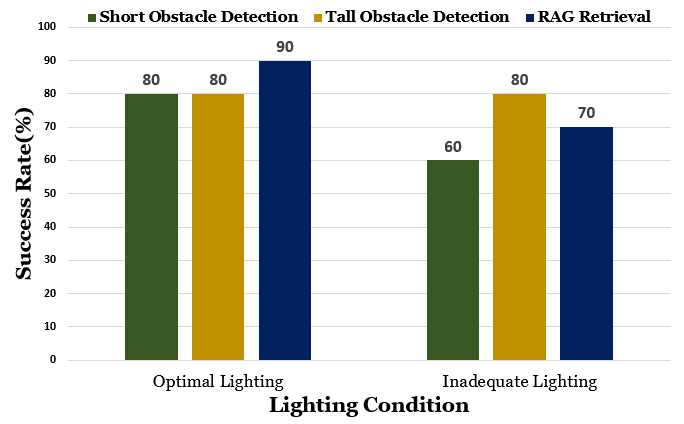}
\caption{Performance of the VLM-RAG system under different lighting conditions.}
\label{fig:VLM-RAG Performance}
\vspace{-5mm}
\end{figure}
\subsection{Evaluation of VLM-RAG Framework}
Fig.~\ref{fig:VLM-RAG Performance} shows that the VLM-RAG system achieves 80\% success in detecting and retrieving short and tall objects under good lighting. In poor lighting, the success rate drops to 60\%, mainly due to difficulty identifying tall obstacles. Despite lighting variations, the system consistently detects obstacles regardless of color or placement.

\section{Conclusion and Future Work}
\textbf{SwarmVLM} introduces a heterogeneous robotic system that integrates the aerial agility of a drone with the ground-level adaptability of a mobile robot, enabling resilient navigation in complex and dynamic environments. Through the use of impedance-based coordination, APF-driven path planning, and vision-language perception via VLM-RAG, the system achieved a navigation success rate of approximately \textbf{92\%} in real-world scenarios. The drone serves as the path leader, while the mobile robot ensures continuous tracking and obstacle avoidance. Moreover, due to its focus on bypassing short obstacles, the mobile robot exhibited a deviation of up to \textbf{50\,cm} from the drone’s trajectory. Furthermore, real-time impedance parameter tuning enhances the system's adaptability to environmental changes. The VLM-RAG module also achieved \textbf{80\%} accuracy in detecting relevant objects and retrieving appropriate control parameters under optimal lighting conditions. Therefore, these combined capabilities enable both agents to adjust their behavior dynamically, ensuring intelligent and dependable operation across varied environments.

In future work, this research can be extended to incorporate multiple drones and mobile robots operating collaboratively. Additionally, integrating advanced computer vision techniques may further enhance real-time impedance parameter tuning by enabling more accurate scene interpretation and contextual awareness.

\balance

\bibliographystyle{IEEEtran}
\bibliography{ref}

\end{document}